\documentclass{article}
\usepackage{spconf,amsmath,graphicx}
\usepackage{amssymb}

\title{AUTOMATED DETECTION OF INDIVIDUAL MICRO-CALCIFICATIONS FROM MAMMOGRAMS USING A MULTI-STAGE CASCADE APPROACH}
 \name{Zhi Lu$^{\ddagger}$ \qquad Gustavo Carneiro$^{\dagger}$ \qquad Neeraj Dhungel$^{\dagger}$  \qquad Andrew P. Bradley$^{\star}$ \sthanks{Supported by the Australian Research Council Discovery Project (DP140102794). Prof. Bradley is an ARC Future Fellow (FT110100623). Z. Lu developed this work while he was at the University of Adelaide.}}
 \address {$^{\dagger}$Australian Centre for Visual Technologies, The University of  Adelaide \\ 
${\ddagger}$  Phenomics and Bioinformatics Research Centre, University of South Australia \\
$^{\star}$ School of ITEE, The University of Queensland }
 \normalsize

\begin{document}
%
\maketitle
\begin{abstract}
In mammography, the efficacy of computer-aided detection methods depends, in part, on the robust localisation of micro-calcifications ($\mu$C). Currently, the most effective methods are based on three steps: 1) detection of individual $\mu$C candidates, 2) clustering of individual $\mu$C candidates, and 3) classification of $\mu$C clusters. Where the second step is motivated both to reduce the number of false positive detections from the first step and on the evidence that malignancy depends on a relatively large number of $\mu$C detections within a certain area. In this paper, we propose a novel approach to $\mu$C detection, consisting of the detection \emph{and} classification of individual $\mu$C candidates, using shape and appearance features, using a cascade of boosting classifiers. The final step in our approach then clusters the remaining individual $\mu$C candidates. The main advantage of this approach lies in its ability to reject a significant number of false positive $\mu$C candidates compared to previously proposed methods. Specifically, on the INbreast dataset, we show that our approach has a true positive rate (TPR) for individual $\mu$Cs of 40\% at one false positive per image (FPI) and a TPR of 80\% at 10 FPI. These results are significantly more accurate than the current state of the art, which has a TPR of less than 1\% at one FPI and a TPR of 10\% at 10 FPI. Our results are competitive with the state of the art at the subsequent stage of detecting clusters of $\mu$Cs.
\end{abstract}
\vspace{-.17cm}
\section{Introduction}
\label{sec:introduction}
\begin{figure}[t]
\begin{center}
\includegraphics[width = 1.4in]{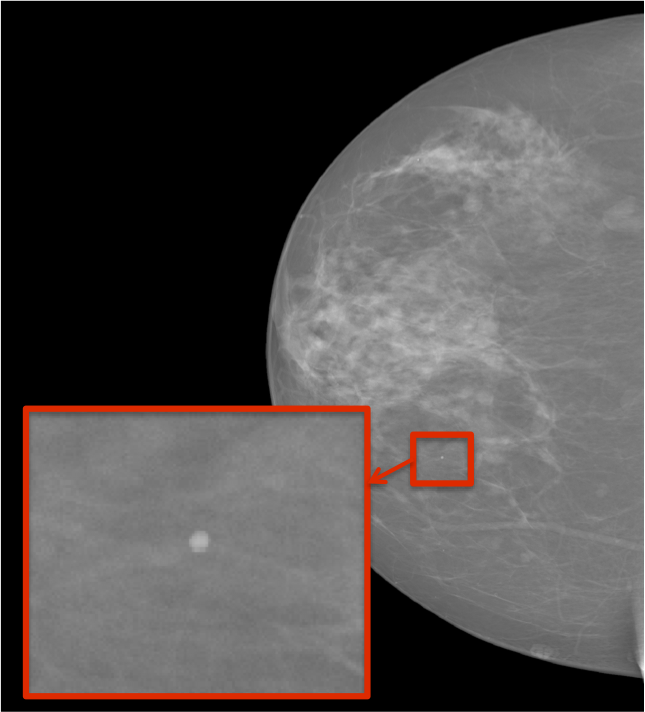} \\
\caption{Appearance of a micro-calcification in a mammogram.}
\label{fig:microcalc_example}
\end{center}
\end{figure}

\vspace{-.17cm}
Breast cancer is the most diagnosed cancer amongst women worldwide, with 23\% of all diagnosed cancers~\cite{jemal2008cancer}.  Breast screening programs aims to detect breast cancer at its early stages, when treatment is generally more effective~\cite{sickles1996breast}.  These programs are usually based on the analysis of mammograms, where one of the main goals is the detection of micro-calcifications ($\mu$C) given that almost half of all breast cancers are associated with $\mu$C~\cite{bick2014mammography}.  As shown in Fig.~\ref{fig:microcalc_example},
$\mu$Cs are represented by tiny calcium deposits that are displayed as small white spots on a mammogram. Their automated localisation by computer-aided detection (CADe) methods has the potential to streamline mammogram analysis and reduce the inter-user variance of $\mu$Cs~\cite{giger2013breast}.  

The current state-of-the-art methods for the automated detection of $\mu$Cs consists of the following standard pipeline: S.1) detection of individual $\mu$C candidates, S.2) clustering of individual $\mu$C candidates based on their geometric distribution, and S.3) classification of those $\mu$C clusters.  
The fact that step S.1 above usually produces a large number of false positive individual $\mu$Cs combined with the evidence that calcification malignancy is correlated with clusters of $\mu$C~\cite{bick2014mammography} has motivated the inclusion of S.2 and S.3.  These last two steps are able to eliminate a large number of isolated false positive $\mu$C detections, but they often fail to reject individual false positive $\mu$Cs within clusters, which can potentially bias the analysis of a mammogram.
In this paper, we propose a novel pipeline comprising the following steps:  P.1) detection of individual $\mu$C candidates, P.2) classification of individual $\mu$C candidates, and P.3) clustering of individual $\mu$C candidates based on their geometric distribution.
Compared to the standard pipeline, steps P.1 and S.1 are the same, step P.2 is new, step P.3 is the same as step S.2, and step S.3 has been removed.
We have two goals with our new approach: 1) a significant reduction of the number of false positive individual $\mu$C detections, especially within true positive clusters, and 2) competitive detection rate of $\mu$C clusters.  
A quantitative analysis of our approach is performed using the publicly available INBreast dataset~\cite{moreira2012inbreast}, where the main results obtained show that our method achieves a true positive detection of individual $\mu$Cs (TPR) of 40\% at 1 false positive detection per image (FPI) and 80\% TPR at 10 FPI, which is significantly more accurate than the current state of the art~\cite{bria2014learning}.  We also show competitive results in terms of the detection of clusters of $\mu$Cs.

\vspace{-.17cm}
\section{Related Work}
\label{sec:related_work}
\vspace{-.17cm}

Automatic detection of $\mu$Cs from mammograms is usually carried out with a combination of image processing and machine learning methods~\cite{shin2015novel}.  Image processing methods~\cite{papadopoulos2008improvement} rely on prior knowledge about the appearance of $\mu$Cs (such as local gradient and intensity). 
However this approach is unlikely to provide a robust characterisation of all variations in the appearance of $\mu$Cs.
Alternatively, machine learning~\cite{el2002support,wei2005relevance} aims to provide a robust characterisation of $\mu$Cs from the information available from an annotated training set. 
In general, with a relatively large and diverse training set, machine learning methods can outperform image processing methods. However, they face some challenges: 1) the selection of an appropriate feature set to be extracted and used by the models, 2) the class imbalance problem that provides a much larger number of negative than positive samples on which to train the model (caused by the significantly smaller area occupied by $\mu$Cs compared with the area filled with normal breast tissue), and 3) the selection of the model to be used.  State-of-the-art machine learning methods address these challenges with the use of a cascade of boosting classifiers that rely on general appearance and shape features~\cite{bria2014learning,oliver2012automatic,shin2015novel}. Therefore, we follow this strategy here.

The design of current state-of-the-art methods consisting of steps S.1-S.3 (defined in Sec.~\ref{sec:introduction}) is based on the clinical importance that clusters of $\mu$Cs have in comparison to individual $\mu$Cs~\cite{bick2014mammography}.  Another motivation is that step S.1 tends to generate a large number of individual false positive $\mu$C detections that need to be subsequently removed.  Hence, by clustering individual $\mu$Cs using proximity and number of detections within a small area, it is possible to reject a large number of isolated false positive detections\cite{bria2014learning}.  However, we observed that within a cluster, false positive detections were still prevalent. This motivated us to propose a new pipeline with the introduction of a classification step between standard steps S.1 and S.2 that filters out individual $\mu$C candidates using their shape and appearance features with a cascade of boosting classifiers~\cite{seiffert2008rusboost}.  Given that this new process is effective for removing false positive $\mu$C detections, we no longer require step S.3 (cluster classification).  

\begin{figure}[t]
\begin{center}
\includegraphics[width = 2.4in]{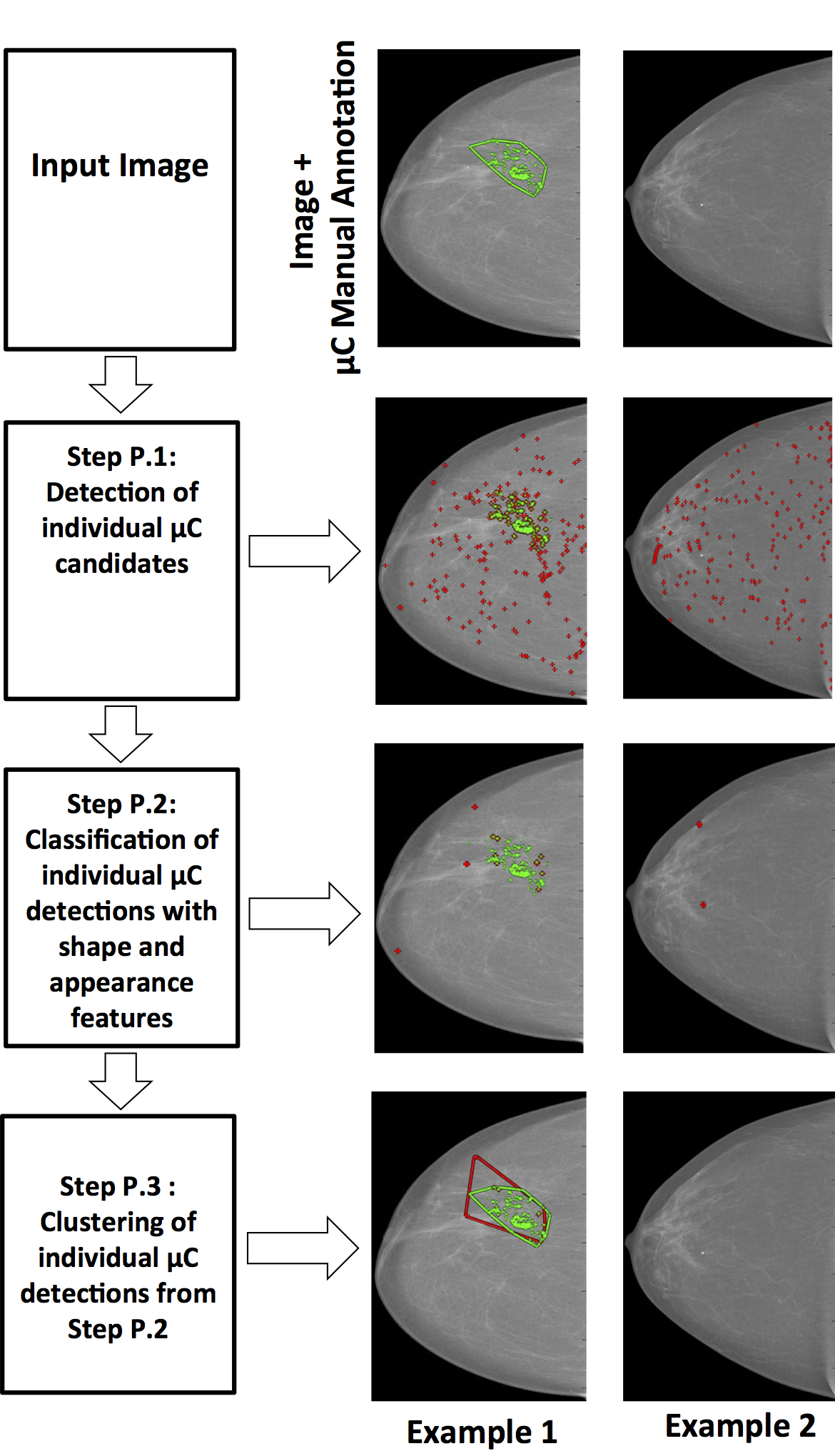} \\
\caption{Pipeline of the proposed methodology with two examples, where the red contours on the mammograms denote the automated detections and green contours represent the manual annotation.}
\label{fig1_workflow}
\end{center}
\end{figure}

\vspace{-.17cm}
\section{Methodology}
\label{sec:method}
\vspace{-.17cm}

The proposed methodology consists of a initial pre-processing step based on quantum noise equalisation~\cite{bria2014learning}, which is followed by three steps: P.1) detection of individual $\mu$C candidates using pixel-based cascade of boosting classifiers~\cite{seiffert2008rusboost} and Haar like features~\cite{bria2014learning}; P.2) classification of the individual $\mu$C detections (from step P.1) using a region-based cascade of boosting classifiers~\cite{seiffert2008rusboost} with appearance and shape features~\cite{varela2006use}; and P.3) clustering of the individual $\mu$Cs detected in step P.2.  Each step is explained below, where we assume the availability of a training set $\mathcal{D} = \{ (\mathbf{x}_i, \mathcal{M}_i) \}_{i=1}^N$, where $\mathbf{x}:\Omega \rightarrow \mathbb R$ denotes the mammogram ($\Omega$ represents the image lattice) and $\mathcal{M} = \{ \mathbf{y}_j \}_{j=1}^J$ represents the set of $\mu$C annotations for image $i$ with $\mathbf{y}:\Omega \rightarrow \{0,1\}$ (i.e., each $\mu$C annotation is a binary map, where pixels of $\mathbf{y}_{i,j}$ labelled with $1$ denote part of the $j^{th}$ $\mu$C of $i^{th}$ image).

{\bf Pre-processing:}  Our pre-processing is based on quantum noise equalisation proposed by Bria et al.~\cite{bria2014learning}, where the source of noise fluctuations in full-field digital mammograms (FFDM) can be described by a Poisson distribution with a standard deviation that can be estimated from the image.  

{\bf Step P.1: Detection of individual $\mu$C candidates:}  This first step consists of a pixel-based classifier~\cite{bria2014learning}, represented by $H(\mathbf{q})$, which estimates the likelihood that the pixel $\mathbf{q} \in \Omega$ represents part of a $\mu$C given the information extracted from a sub-window of size $M \times M$ around the pixel.  This classifier is represented by a cascade of boosting classifiers, where a pixel $\mathbf{q}$ is accepted to be part of a $\mu$C if it is positively classified by all stages of the cascade.  In this cascade classifier~\cite{viola2001rapid}, the detection rate $D$ and false positive rate $F$ of a cascade with $S$ stages are computed with $D = \prod_{s=1}^S d_s$ and $F = \prod_{s=1}^S f_s$, where $d_s$ and $f_s$ represent the detection and false positive rates of stage $s$.  Therefore, if $d_s = 0.99$ and $f_s = 0.3$ and $S=5$, then $D = 0.951$ and $F=0.002$.


The training of the classifier at each cascade stage $s \in \{1,...,S\}$, denoted by $H_s(\mathbf{p})$, uses a set of positive samples $\mathcal{P}_s$ and negative samples $\mathcal{N}_s$, where each sample $\mathbf{x}_{M\times M}(\mathbf{q}) \in \mathcal{P}_s$ consists of a sub-image of $\mathbf{x}$ of size $M \times M$ centred at position $\mathbf{q}$, such that one of the $J$ $\mu$C annotations contains $\mathbf{y}(\mathbf{q})=1$ (a negative sample is similarly defined with $\mathbf{y}(\mathbf{q})=0$).  The main issue in training such classifier is the fact that $|\mathcal{N}_s| >> |\mathcal{P}_s|$, and this is solved by under-sampling the negative set, such that the proportion $|\mathcal{N}_s|/|\mathcal{P}_s|$ is constant over the training of each cascade stage.  The classifier utilised in this work is the RUSBoost~\cite{seiffert2008rusboost}, which is designed to deal with such class imbalance with this under-sampling procedure. Finally, the feature set used is the Haar-like features~\cite{viola2001rapid}, which are efficiently computed using integral image~\cite{viola2001rapid} (note that we use a set of $1,697$ features instead of the original $14,709$ features from~\cite{bria2014learning} as this smaller set is faster to train and we did not notice a significant difference in the results).  The final part of this step consists of finding the connected components of the pixel-based classification to form the $\mu$C candidates, where connected components that have width and length larger than $1$ mm are removed because they represent macro-calcifications that are not to be processed further~\cite{bria2014learning}.  The step P.1 is defined by:
\begin{equation}
\{ \tilde{\mathbf{y}}_k, \mathbf{d}_k \}_{k=1}^K  = f_1(\mathbf{x},\theta_1),
\label{eq:f_1}
\end{equation}
where $\tilde{\mathbf{y}}_k$ denotes a binary map of the $k^{th}$ $\mu$C candidate, $\theta_1$ is the classifier parameter set, and $ \mathbf{d}_k \in \mathbb R^4$ represents the  top-left and bottom-right corner coordinates of the bounding box of this detection.

{\bf Step P.2: Classification of individual $\mu$C detections with shape and appearance features:}  The contribution of this paper consists of this individual $\mu$C classification step, where we extract a large set of shape and appearance features~\cite{varela2006use} from each $\mu$C candidate in (\ref{eq:f_1}), and use a second cascade of RUSBoost~\cite{seiffert2008rusboost} classifiers to further eliminate false positive $\mu$C detections.  These features are extracted with:
\begin{equation}
 \mathbf{z} = g(\mathbf{x},\mathbf{d},\mathbf{y}).
\label{eq:features}
\end{equation}
A set of 11 shape features are calculated from $\mathbf{y}$ in (\ref{eq:features}), which describe the following geometric information: area, perimeter, ratio of perimeter to area, rectangularity, circularity, and etc. Another set of 27 appearance features in (\ref{eq:features}) are calculated from the sub-image of $\mathbf{x}$ limited by the bounding box $\mathbf{d}$, consisting of information (energy, correlation, entropy, inertia, and etc.) extracted from the spatial grey level dependence (SGLD) matrix~\cite{varela2006use, dhungel2015automated}. In addition, we compute the $1,697$ Haar-like features of step P.1 and the local binary pattern (LBP)~\cite{ojala1994performance} from the sub-image $\mathbf{x}$ limited by the bounding box $\mathbf{d}$.  The step P.2 is defined by:
\begin{equation}
\{ \tilde{\mathbf{y}}_l, \mathbf{d}_l \}_{l=1}^L  = f_2(\mathbf{x},\{ \tilde{\mathbf{y}}_k, \mathbf{d}_k \}_{k=1}^K,\theta_2),
\label{eq:f_2}
\end{equation}
which selects a subset of the detections from step P.1, with $L \leq K$, where $\theta_2$ is the parameter set of the classifier.

{\bf Step P.3: Clustering of individual $\mu$C detections:}  The clustering of the $\mu$C detections $\{ \tilde{\mathbf{y}}_l, \mathbf{d}_l \}_{l=1}^L$ from step P.2 is based on the following algorithm~\cite{bria2014learning}: 1) construction of a weighted graph formed by nodes represented by the centroid of the detected $\mu$Cs, and edges that connect nodes that are closer than $10$ mm; and 2) estimation of clusters from the connected components of this graph, where clusters with fewer than 3 $\mu$Cs are rejected.  Step P.3 is defined by:
\begin{equation}
\mathcal{C}  = f_3(\mathbf{x},\{ \tilde{\mathbf{y}}_l, \mathbf{d}_l \}_{l=1}^L,\theta_3),
\label{eq:f_3}
\end{equation}
where $\theta_3$ is the parameter set of the classifier, and $\mathcal{C}$ represents the set of clusters, where each element of this set is formed by a graph computed from a subset of $\{ \tilde{\mathbf{y}}_l, \mathbf{d}_l \}_{l=1}^L $ from step P.2.

\begin{figure}[t]
\begin{center}
\includegraphics[width = 3.5in]{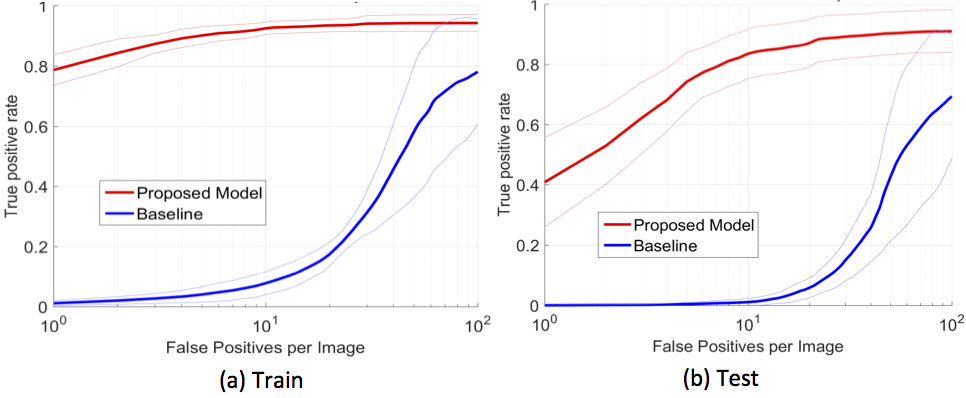} \\
\caption{FROC curves of the {\bf individual $\mu$Cs detections} for our methodology (red) and the baseline (blue)~\cite{bria2014learning}.}
\label{fig:froc_individual_mc}
\end{center}
\end{figure}

\begin{figure}[t]
\begin{center}
\hspace{-.15in} \includegraphics[width = 3.5in]{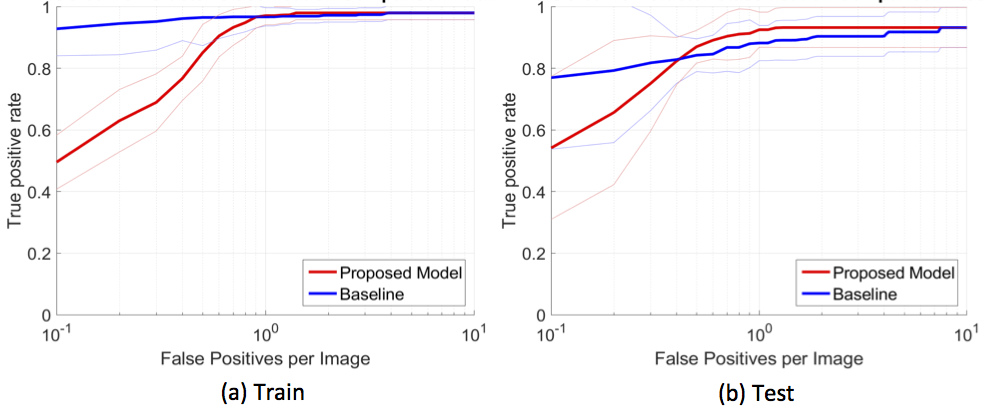} \\
\caption{FROC curves of the {\bf cluster $\mu$Cs detections} for our methodology (red) and the baseline (blue)~\cite{bria2014learning}.}
\label{fig:froc_cluster_mc}
\end{center}
\end{figure}

\begin{figure}[t]
\begin{center}
\hspace{-.15in} \includegraphics[width = 3.5in]{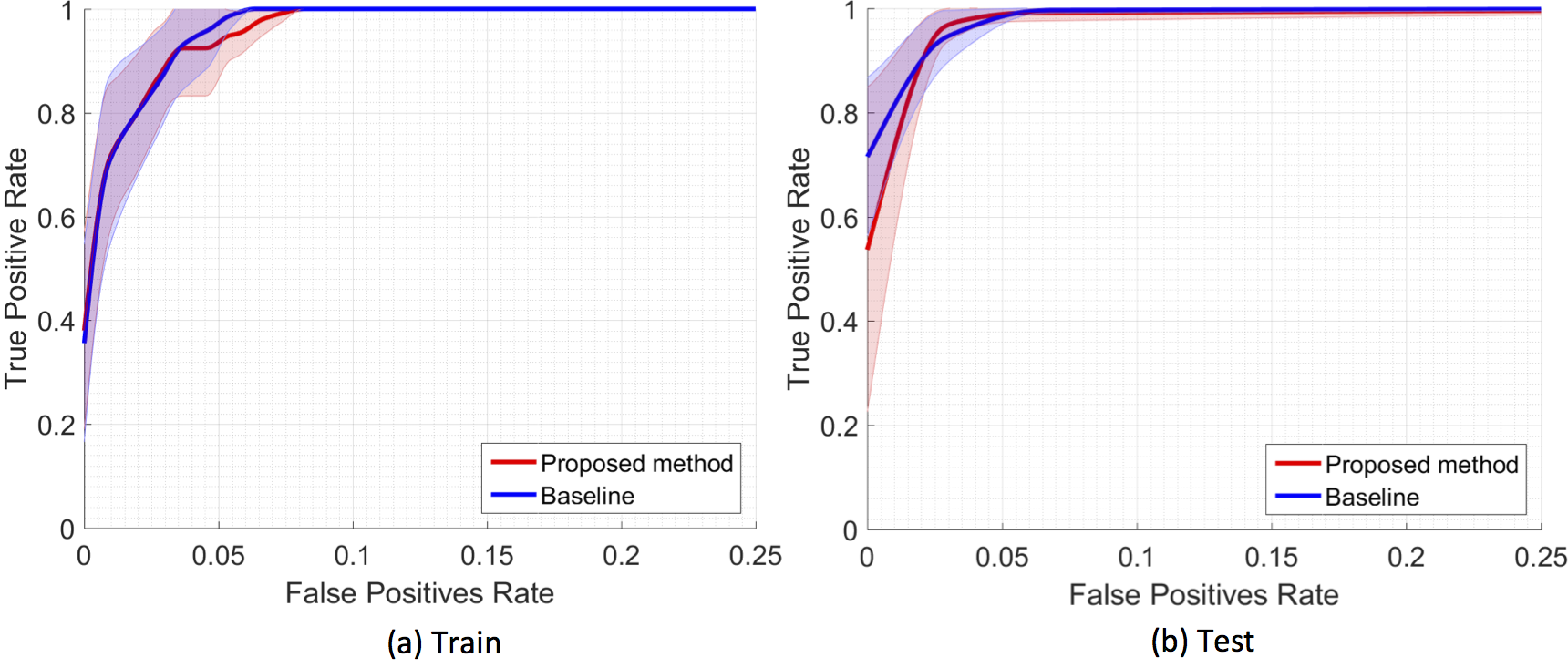} \\
\caption{ROC curves of the {\bf case-based detections of clusters of $\mu$Cs} for our methodology (red) and the baseline (blue)~\cite{bria2014learning}.}
\label{fig:roc_case_mc}
\end{center}
\end{figure}

\vspace{-.17cm}
\section{Materials and Methods}
\label{sec:experiments}
\vspace{-.17cm}

The experiments use the INBreast dataset~\cite{moreira2012inbreast}, which contains 115 cases with 410 images, where 19 cases have no findings, 68 cases have benign findings and 28 cases have malignant findings (note that findings include $\mu$C and masses), where 6,880 individual $\mu$Cs have been identified by two radiologists.
The experiments are performed using this dataset for the following reasons:
it is a public domain (allowing direct comparison with other methods) full-field digital dataset where the individual manual $\mu$C annotations are both precise and reliable.  In order to evaluate the detection of $\mu$C clusters, we produce the annotation of $\mu$C clusters using step P.3 of Sec.~\ref{sec:method} from the individual $\mu$C manual annotations.
We perform a quantitative evaluation of the individual $\mu$C detection and cluster of $\mu$Cs detection by randomly dividing the 115 INBreast cases into five cross-validation folds with 60\% of cases for training, 20\% for validation and 20\% for testing. We show the mean and standard deviation of performance on the both the train and test sets (note that we interpolate the ROC and FROC curves for each fold at fixed FPR values in order to plot the mean and standard deviation error bars).

The training of the classifier in step P.1 (Sec.~\ref{sec:method}) uses sub-images of size $M \times M$, with $M=12$ pixels (slightly less than $1 \times 1$ mm), which is approximately the maximum size of the $\mu$C of interest (i.e., between $0.1$ and $1$ mm~\cite{bria2014learning}).  For step P.1, we train five cascade stages, where the ratio between the sizes of the negative and positive sets (step P.1 of Sec.~\ref{sec:method}) is fixed to be $1 \times 10^6$ for all these stages.  The RUSBoost classifier has 2, 3, 5, 12, and 40 weak classifiers~\cite{bria2014learning} in each of the stages, where the detection rate is $d_s = 0.99$ and false positive rate is $f_s = 0.30$~\cite{bria2014learning}.
For step P.2, the connected components forming the $\mu$C candidates are also resized to a fixed size of $12\times 12$ patches using bicubic interpolation, from which the shape and appearance features are calculated.  The classifier in step P.2 is a single RUSBoost classifier with 1000 weak learners.  Model selection was performed on the validation set extracted from the training set of each cross validation fold.  
We compare our method with the baseline approach by Bria et al.~\cite{bria2014learning}, which consists of a methodology that contains the same pre-processing, followed by steps P.1 and P.3, and a final step that classifies the clusters (this baseline is the standard baseline described in Sec.~\ref{sec:introduction}).  
The implementation of this cluster classification is based on a cascade of RUSBoost~\cite{seiffert2008rusboost} classifiers that use the mean, standard deviation, minimum and maximum values of 35 shape and appearance features from individual $\mu$Cs and 5 topological cluster features (i.e., a total of $4 \times 35 + 5$ features = $145$ features).

\begin{figure}[tb]
\begin{center}
\includegraphics[width = 8cm]{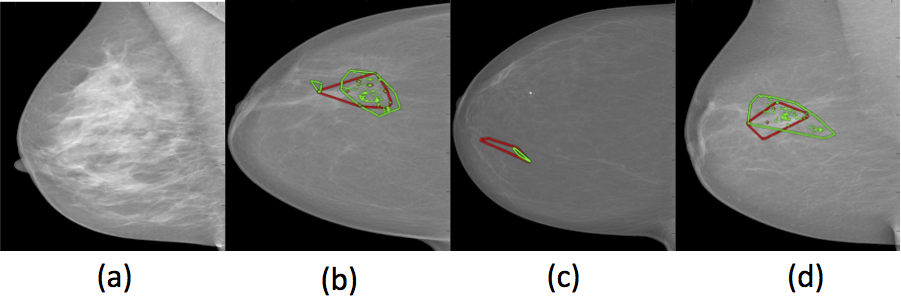} 
\caption{Examples of $\mu$C detection produced by our method (green = manual annotation, red = automated detection).}
\label{fig:visual_inspection_examples}
\end{center}
\end{figure}

The quantitative evaluation of individual $\mu$Cs detections and clusters of $\mu$Cs is based on free-response receiver operating characteristic (FROC) curve that measures the true positive rate (TPR). An individual $\mu$Cs is considered as true positive detection if it has an overlap of at least $0.5$ with one of the manual annotations 
and a false positive detection if it has an overlap less than $0.5$ with any manual annotation.   Similarly, clusters of $\mu$C detections are regarded as a true positive if they overlap with a manually annotated cluster and where they have at least two individual $\mu$Cs in common~\cite{jing2010detection,bria2014learning}. Finally, the case-based ROC curve evaluates the performance of the method in terms of finding  $\mu$C clusters independently of whether they are in the correct location. Here, a true positive is defined as a cluster detection in a case that has at least one manually annotated cluster of $\mu$Cs, and a false positive is a detection in case that has no manually annotated cluster. Fig.~\ref{fig:froc_individual_mc} shows the FROC curves of individual $\mu$Cs detections on training and testing sets for our methodology (red) and the baseline (blue)~\cite{bria2014learning} (note that we crop the FPI at $10^2$, but TPR continues to increase for all curves). In Fig.~\ref{fig:froc_cluster_mc}, we show the results of our methodology on cluster of $\mu$Cs detection (red) and the baseline result (blue)~\cite{bria2014learning}. Similarly, Fig.~\ref{fig:roc_case_mc} shows the ROC curves of the case-based detections of clusters of $\mu$Cs on training and testing sets for our methodology (red) and the baseline (blue)~\cite{bria2014learning}.  Finally, we show some visual examples of the final detections of our method in Fig.~\ref{fig:visual_inspection_examples}.

\vspace{-.17cm}
\section{Discussion and Conclusions}
\label{sec:discussion}
\vspace{-.17cm}

The results from the Fig.~\ref{fig:froc_individual_mc} show that our approach is significantly more effective at the detection of individual $\mu$Cs compared to the baseline~\cite{bria2014learning}. It is also interesting to note from Fig.~\ref{fig:froc_cluster_mc} and Fig.~\ref{fig:roc_case_mc} that our approach is competitive with the baseline in terms of cluster detection and case-based performance (note that the the results in those figures agree with the published results by Bria at al.~\cite{bria2014learning}, even though we use a different dataset). %
This apparent discrepancy in results is explained by the large number of individual false positive $\mu$C detections that are preserved within true positive clusters of $\mu$C detections by the baseline approach~\cite{bria2014learning}.  Our method is able to eliminate a significant number of these false positives and thus provide a more reliable result on which to perform further assessment of the mammogram. Finally, Fig.~\ref{fig:visual_inspection_examples}(a) shows that our proposed methodology is robust to normal mammograms, while Fig.~\ref{fig:visual_inspection_examples}(b-d) displays visually accurate detection of the individual as well as clusters of $\mu$Cs.

In this paper we propose a new $\mu$C detection pipeline that introduces a step that effectively filters out individual false positive $\mu$C detections using shape and appearance features in a cascade of boosting classifiers.  We empirically show that our method displays a significantly more effective detection of individual $\mu$Cs compared to the current state-of-the-art approach~\cite{bria2014learning}. This has the potential to improve the mammogram analysis in breast screening programs.


\bibliographystyle{IEEEbib}
\bibliography{refs}

\end{document}